\documentclass{article}
\usepackage{spconf,amsmath,graphicx}
\usepackage{todonotes}
\usepackage{multirow}
\usepackage{graphicx}
\usepackage[normalem]{ulem}
\useunder{\uline}{\ul}{}
\usepackage{hyperref}
\usepackage{booktabs}

\usepackage{makecell}
\usepackage{rotating}

\usepackage{subfiles}

\input{tikzstuff}

\title{SurvivalNet: Predicting patient survival from diffusion weighted magnetic resonance images using cascaded fully convolutional and 3D convolutional neural networks}
\makeatletter
\def\@name{
Patrick Ferdinand Christ
$^{1,\star}$ \qquad Florian Ettlinger$^{1,\star}$ \qquad Georgios Kaissis$^{1,\dagger}$\\

\qquad Sebastian Schlecht$^{\star}$ \qquad Freba Ahmaddy$^{\dagger}$ 
 \qquad Felix Gr\"un$^{\star}$ \qquad Alexander Valentinitsch$^{\dagger}$ 
\\
\qquad Seyed-Ahmad Ahmadi$^{+}$ \qquad Rickmer Braren$^{2,\dagger}$ \qquad Bjoern Menze$^{2,\star}$
}
\makeatother
 \address{ \small $^{\star}$ Technische Universit\"at M\"unchen, Image-Based Biomedical Modeling Group, Arccisstrasse 21, 80333 Munich \\
 \small   $^{\dagger}$Technische Universit\"at M\"unchen, Institute for diagnostic and interventional Radiology, Ismaninger Str. 22, 81675 Munich \\
  \small   $^{+}$ Ludwig-Maximilians-Universit\"at,
German Center for Vertigo and Balance Disorders,
Feodor-Lynen-Stra{\ss}e 19,
81377 Munich }

\begin{document}

\maketitle

\begin{abstract}
Automatic non-invasive assessment of hepatocellular carcinoma (HCC) malignancy has the potential to substantially enhance tumor treatment strategies for HCC patients. In this work we present a novel framework to automatically characterize the malignancy of HCC lesions from DWI images.

We predict HCC malignancy in two steps: As a first step we automatically segment HCC tumor lesions using cascaded fully convolutional neural networks (CFCN). A 3D neural network (SurvivalNet) then predicts the HCC lesions' malignancy from the HCC tumor segmentation. We formulate this task as a classification problem with classes being ``low risk" and ``high risk" represented by longer or shorter survival times than the median survival. We evaluated our method on DWI of 31 HCC patients. Our proposed framework achieves an end-to-end accuracy of 65\% with a Dice score for the automatic lesion segmentation of 69\% and an accuracy of 68\% for tumor malignancy classification based on expert annotations. We compared the SurvivalNet to classical handcrafted features such as Histogram and Haralick and show experimentally that SurvivalNet outperforms the handcrafted features in HCC malignancy classification. End-to-end assessment of tumor malignancy based on our proposed fully automatic framework corresponds to assessment based on expert annotations with high significance ($p>0.95$).  
\end{abstract}

\section{Introduction}
\footnotetext[1]{Authors contributed equally}
\footnotetext[2]{Corresponding authors: rbraren@tum.de and bjoern.menze@tum.de}
\subsection{Motivation}
Hepatocellular carcinoma (HCC) presents the sixth most common cancer and the third most common cause of cancer-related deaths worldwide \cite{ferlay2010estimates}. HCC comprises a genetically and molecularly highly heterogeneous group of cancers that commonly arise in a chronically damaged liver. Importantly, HCC subtypes differ significantly in clinical outcome. The stepwise transformation to HCC is accompanied by major changes in tissue architecture including an increase in cellularity and a switch in vascular supply (i.e. arterialization). These differences provide the basis for the non-invasive detection of HCC \cite{european2012easl}. 
In particular, diffusion weighted-magnetic resonance imaging (DW-MRI) detects differences in random Brownian motion, which is commonly reduced in highly cellular HCC due to an increase in cell membranes and macromolecules. The apparent diffusion coefficient (ADC) parameter value, which can be derived from two DW-MRI scans, quantifies this effect. DW-MRI imaging techniques provide a high level of sensitivity and specificity for tumor detection, the distinction of tumor subtypes requires the identification of more subtle differences. Computer aided analysis techniques allow medical image feature extraction far beyond the capabilities of the human eye and thus hold the potential for an imaging based differentiation of tumor subtypes. Non-invasive differentiation of tumor subtypes in HCC would enable pre-therapeutic patient stratification and the systematic testing of novel therapeutic strategies.

\begin{figure*}[t]
\includegraphics[width=\textwidth]{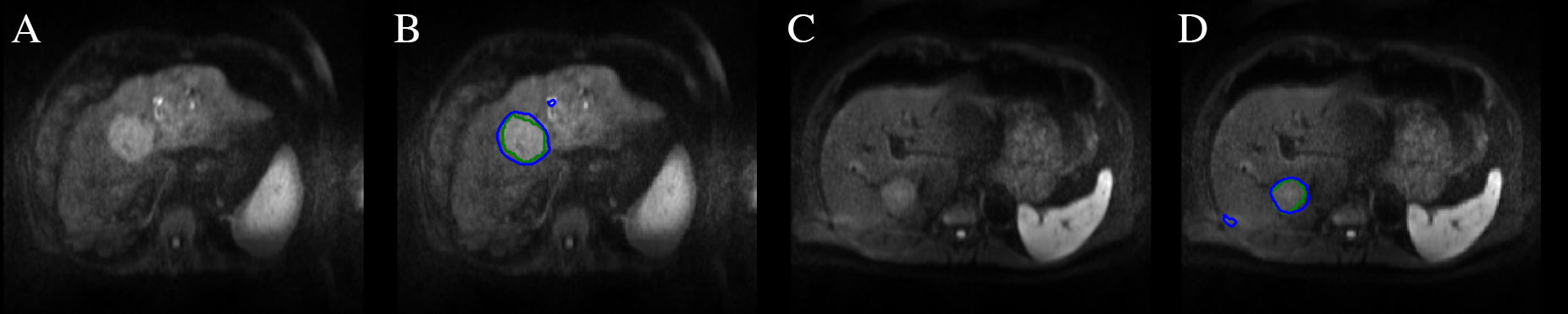}
\caption{Results of the automatic HCC tumor segmentation in DW-MRI: A and C show the DW-MRI slice. B and D show the ground truth label of HCC in green and the automatic segmentation using CFCN in blue. The automatic segmentation algorithms successfully segments the HCC tumor in both cases. Only two small false-positive regions and a slight inaccuracy around the edges of the tumors are visible leading to a Dice overlap score of 85 \% for B and 83 \% for D. The patient in A/B belongs to class ``high risk" whereas C/D belongs to class ``low risk". Yet, only subtle overall differences in appearance between the tumors are visible. }
\label{fig:seg}
\end{figure*}
\subsection{Related Works}
Heid et al. (2016) have recently established a close relationship between the regional DW-MRI derived apparent diffusion coefficient (ADC) parameter value and distinct subtypes in pancreatic ductal adenocarcinoma (PDAC) \cite{Heidclincanres}.  Computer aided extraction of image features for tumor subtyping has previously been reported for several tumor entities. Prior work focused mostly on hand-crafted feature extraction such as histogram features \cite{Lee2014}, Gabor and Haralick features \cite{Yao2016,zhu}, and grey level run length based features \cite{song} to predict survival times for diverse tumor entities and image modalities. Recent works leveraged the discriminative power of the apparent diffusion coefficient (ADC) by extracting texture features for survival or malignancy characterization \cite{Reda2016,Shehata2016}. Zhou et al. (2016) proposed a method to characterize malignancy of HCC in contrast enhanced MRI by extracting histogram and texture based features such as grey-level co-occurrence and run length (GLRL and GLCM) of HCC lesions \cite{zhou2016malignancy}. However, their method required manual segmentation of the lesions beforehand.
\subsection{Contribution}
In comparison to prior work, we developed a method to predict HCC survival from DW-MRI volumes using automatic segmentations of tumors. Our contribution in this work is three-fold. First, we developed an automatic method to detect and segment HCC tumor lesions in DW-MRI data. Second, we found and analyzed quantitative biomarkers using handcrafted and CNN-based features to predict patient survival. Third, we experimentally demonstrated a fully automatic method to predict long/short survival of HCC patients from DW-MRI images.
\begin{figure*}
\includegraphics[width=\textwidth]{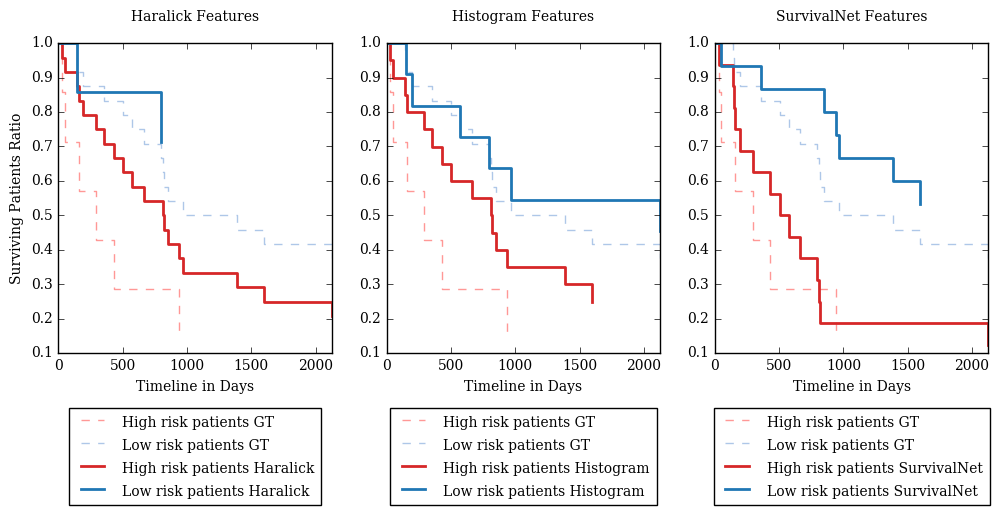}
\caption{Kaplan-Meier Survival Analysis for Haralick Features (left), Histogram Features (middle) and 3D SurvivalNet CNN (right). The SurvivalNet is able to split the HCC patients into high risk i.e. short survival and low risk i.e. long survival. In contrast, classical handcrafted features such as histogram and Haralick do not predict the patient survival correctly. GT stands for ground truth.}
\label{fig:kaplan}
\end{figure*}

\section{Methods}
\begin{figure}
\begin{tikzpicture}[box/.style={line width=2pt,draw=tabBlue,rounded corners=8pt,minimum  width=\columnwidth,align=center,fill=tabBlueLight,inner sep=3mm,node distance=6mm},arrow/.style={line width=28mm,arrows={-Triangle[angle=150:15mm]},draw=tabGray,fill=tabGray}]
\node[box] (mri) {\textbf{MRI-Data}\\\small DWI + ADC Volumes};
\node[box,below=of mri] (cnn) {\textbf{Automatic Lesion Segmentation}\\\small Cascaded Fully Convolutional Network};
\node[box,below=of cnn] (features) {\textbf{Feature Extraction}\\\small Histogram, Texture and 3D CNN Features};
\node[box,below=of features] (classifier) {\textbf{Survival Prediction}\\\small Softmax-Classifier};

\draw [arrow] (mri) edge (cnn);
\draw [arrow] (cnn) edge (features);
\draw [arrow] (features) edge (classifier);


\end{tikzpicture}
\caption{Framework for lesion segmentation and survival prediction}
\label{fig:pipeline}
\end{figure}
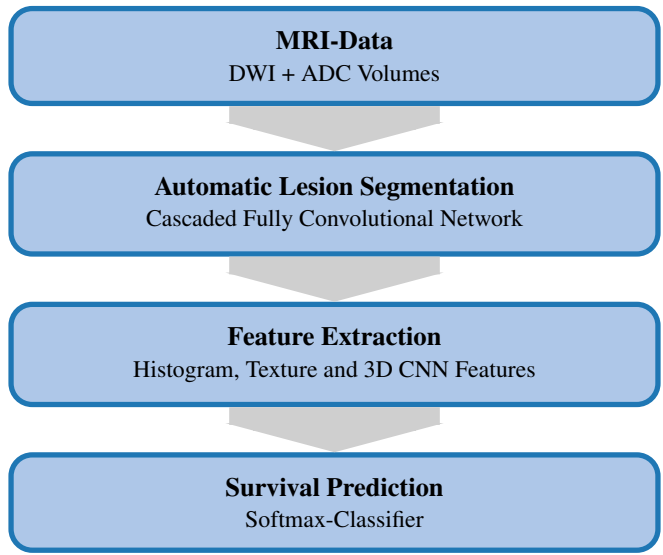

Our proposed framework to fully automatically predict short/long survival from DW-MRI of HCC patients is depicted in figure \ref{fig:pipeline}.
\subsection{Dataset}
31 Patients underwent clinical assessment and MR imaging for the primary diagnosis of HCC. Barcelona Clinic Liver Cancer Classification was used to assess the clinical stage of the disease. Patients with a history of prior malignancy were excluded. No data with insufficient quality due to breathing artifacts, excessive banding or distortion, diffuse tumor growth or non-detectability of the lesions in the DW-MRI sequences was included in the dataset. Imaging was performed using a 1.5 T clinical MRI scanner (Avanto, Siemens) with a standard imaging protocol including axial and coronal T2w, axial T1w images before and after application of Gadolinium-DTPA contrast agent (Jenapharm Magnograf ® 0.5 mmol/ml per manufacturer’s instructions). Post-contrast T1w images were acquired in the early, mid and late arterial phases as well as in the portal venous phase. Diffusion weighted imaging was performed using a slice thickness of 5mm and a matrix size of 192 by 192. Institutional review board approval was obtained for this retrospective study.

\subsection{Automatic Segmentation}
To automatically detect and segment tumor lesions we applied a cascaded fully convolutional neural network to segment in step 1 the liver and in a step 2 the tumor lesions from a liver ROI volume \cite{Christ2016}. We used the DW-MRI as input to the FCN architecture proposed by Ronneberger et al. (2015) \cite{Unet}. We fine-tuned our networks using the liver and liver tumor model provided by Christ et al. (2016) \cite{Christ2016} and applied a 5-fold cross-validation. 
Tumor margins were identified in the early arterial phase and in DWI images (b=600). Manual segmentation was performed by an experienced radiologist using the software TurtleSeg\textsuperscript{\textregistered} and reviewed by two expert radiologists.

\subsection{Survival Prediction}

To predict the survival rate of HCC tumor patients we calculated different features using the detected and segmented tumor lesions applied in the ADC image sequences. We calculated handcrafted features and features trained end-to-end by a 3D Convolutional Neural Network (SurvivalNet).
\subsubsection{Handcrafted Features}
ADC value histograms were generated from the regions of the ADC map corresponding to the tumor ROI in the b=600 image. Histogram descriptors were obtained including mean, median, kurtosis and skewness. In addition, we extracted features representing ADC texture by calculating 3D Haralick statistics of the grey-level co-occurrence matrix \cite{haralick79}. We trained a k-nearest-neighbour classifier with k=4 and validated the results using 10-fold cross-validation.
\subsubsection{SurvivalNet: 3D Convolutional Neural Network}
\begin{figure}[h!]
\begin{tikzpicture}[
box/.style={line width=1pt,draw=tabBlue,fill=tabBlueLight,
  minimum width=4cm,minimum height=0.9cm, node distance=4mm},
  orangeBox/.style={draw=tabOrange,fill=tabOrangeLight},
  greenBox/.style={draw=tabGreen,fill=tabGreenLight},
  redBox/.style={draw=tabRed,fill=tabRedLight},
  arrow/.style={line width=28mm,arrows={-Triangle[angle=150:15mm]},draw=tabGray,fill=tabGray},
  dummybox/.style={minimum width=4cm,minimum height=0.9cm, node distance=4mm}]

\node[dummybox] (input) {dummy};
\node[dummybox,below=of input] (conv1) {dummy};
\node[dummybox,below=of conv1] (maxpool1) {dummy};
\node[dummybox,below=of maxpool1] (conv2) {dummy};
\node[dummybox,below=of conv2] (maxpool2) {dummy};
\node[dummybox,below=of maxpool2] (fc1) {dummy};
\node[dummybox,below=of fc1] (fc2) {dummy};

\node[below=of fc2,inner sep=3mm] (labels) {High Risk vs. Low Risk};

\draw [arrow] (input) edge (labels);

\node[parallelepiped,box] (input) {\textbf{Input}};
\node[parallelepiped,box, orangeBox, below=of input] (conv1) {\textbf{3D Convolution}};
\node[parallelepiped,box, greenBox, below=of conv1] (maxpool1) {\textbf{3D Maxpool}};
\node[parallelepiped,box, orangeBox, below=of maxpool1] (conv2) {\textbf{3D Convolution}};
\node[parallelepiped,box, greenBox, below=of conv2] (maxpool2) {\textbf{3D Maxpool}};
\node[box, redBox, below=of maxpool2] (fc1) {\textbf{Fully Connected}};
\node[box, redBox, below=of fc1] (fc2) {\textbf{Fully Connected}};

\node[right=of input] (input_dim) {64x64x64 voxels};
\node[right=of conv1,text width=3cm] (conv1_dim) {50 filters\\ 3x3x3 kernels};
\node[right=of maxpool1,text width=3cm] (maxpool1_dim) {2x2x2 pool size};
\node[right=of conv2,text width=3cm] (conv2_dim) {50 filters\\ 3x3x3 kernels};
\node[right=of maxpool2,text width=3cm] (maxpool2_dim) {2x2x2 pool size};
\node[right=of fc1,text width=3cm] (fc1_dim) {500 nodes};
\node[right=of fc2,text width=3cm] (fc2_dim) {1 output};


\end{tikzpicture}
\caption{The SurvivalNet is a 3D CNN that consists of two blocks of 3D convolution followed by 3D maxpooling and finally two fully connected layers.}
\label{fig:survivalnet}
\end{figure}
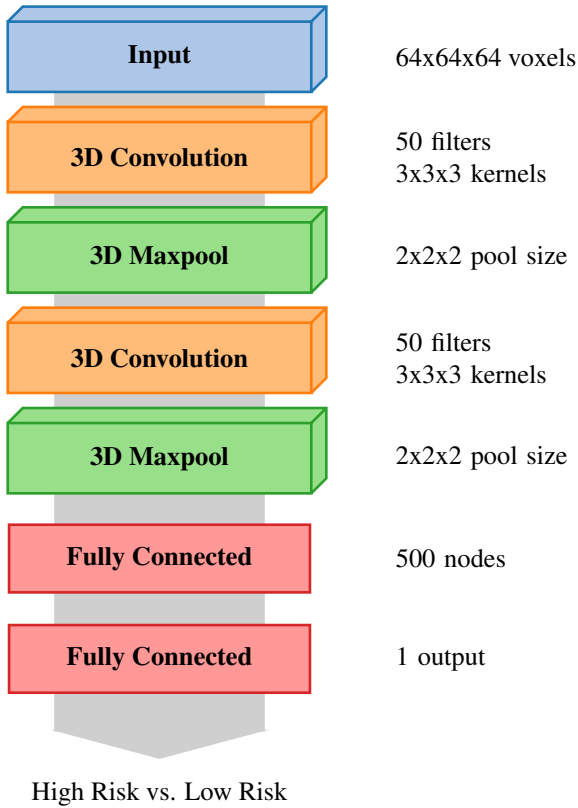

Finally, we trained a 3D CNN to predict the survival rate in an end-to-end fashion. The SurvivalNet consists of two stacks of 3D convolution and max pooling layers, followed by 2 fully-connected layers. The 3D convolutions have 50 kernels with a kernel size of 3x3x3 pixels and a 3D spatial dropout with $p=0.3$. The first fully-connected layer has 500 neurons. Figure \ref{fig:survivalnet} shows the SurvivalNet architecture. We trained the SurvivalNet from scratch using the Adadelta gradient-descent algorithm \cite{adadelta} at a learning rate of $0.5$, $\rho = 0.95$ and $\epsilon = 1 \cdot10^{-8} $. We employed no data-augmentation.

Table \ref{tab:feat} shows the performance of the handcrafted features and SurvivalNet.

\section{Results}
\subsection{Qualitative}
The qualitative results of the automatic segmentation are depicted in figure \ref{fig:seg}. The complex and heterogeneous shape of the tumor lesions was detected and segmented in both images using our automatic segmentation algorithm. The trained model achieves a Dice overlap score of 85\% in both images. The segmentation reaches a high level of specificity by classifying all lesion pixels in the image as lesions. Small false positive outliers within the liver reduce the overall accuracy and Dice score.


\subsection{Quantitative}

\begin{table}[]
\centering
\caption{Quantitative tumor segmentation results}
\label{tab:seg}
\resizebox{\columnwidth}{!}{%
\begin{tabular}{@{}lcccccc@{}}
\toprule
Method            & Sensitivity      & Precision  & TNR         & RVD      & \multicolumn{1}{c}{Dice}     \\
                    & {[}\%{]} & {[}\%{]} & {[}\%{]} & {[}\%{]} &{[}\%{]} \\ \midrule

Cascaded FCN on DWI   &  91.1        &  70.0   & 99.6     &      52.1    &      69.7              
                     \\ \bottomrule
\end{tabular}
}
\end{table}

\renewcommand\theadfont{\normalsize}

\begin{table}[]
\centering
\settowidth\rotheadsize{\theadfont Automatic Tumor}
\renewcommand\theadfont{\normalsize}
\caption{Survival prediction results using both manual tumor segmentations and the output of the automatic tumor segmentation as inputs for the survival prediction classifier}
\label{tab:feat}
\resizebox{\columnwidth}{!}{%
\begin{tabular}{@{}clccccc@{}}
\toprule
                                              & \multirow{2}{*}{Features}           & ACC & Precision & Sensitivity &  F1-Score \\ 
                                              & & {[}\%{]} & {[}\%{]} & {[}\%{]} &{[}\%{]} \\
                                               \midrule
\multirow{3}{*}{Manual Tumor Seg.}          & SurvivalNet CNN                &  68   &  69   &  68   &  65   &    \\
                                              & Histogram Features & 61    &   62  & 61   &  60   &     \\
                                    
                                              & Texture Features: 3D Haralick   &  61   &   65  &  61   &58          \\ \midrule
\multirow{3}{*}{Automatic Tumor Seg.} & SurvivalNet CNN                & 65    & 64    & 65    &   64       \\
                                              & Histogram Features & 58    &  59   &    58 &   56  &    \\

                                              & Texture Features: 3D Haralick   & 61    & 62    & 62    &60          \\ \bottomrule
\end{tabular}%
}
\end{table}

The quantitative results for our automatic segmentation method are shown in table \ref{tab:seg}. Our automatic HCC lesion segmentation algorithm achieves a Dice overlap score of 69.7\% trained on DW-MRI images. The trained model is highly sensitive in recognizing HCC lesions with a Sensitivity of 91.1\%, i.e. only few false negative errors occur.

Table \ref{tab:feat} shows the quantitative results of our proposed automatic survival prediction framework. Figure \ref{fig:kaplan} shows a Kaplan-Meier plot of the survival prediction results. SurvivalNet achieves higher scores on both manual and automatic segmentation compared to handcrafted features. SurvivalNet trained on manual segmentations achieves an accuracy of 68\% with a Precision and Sensitivity of 69\% and 68\% respectively. Furthermore, SurvivalNet accomplishes a classification accuracy of 65\% at a Precision and Sensitivity of 64\% and 65\% when trained using our automatic tumor segmentation in a fully automatic fashion.

As a final experiment, we calculated a paired Wilcoxon signed-rank test with H0: the output posterior class probabilities of SurvivalNet with manual and automatic segmentation belong to the same distribution. At $p>0.953$, we found H0 to be confirmed, i.e. SurvivalNet produces the same results with automatic segmentation or manual segmentation.

\section{Conclusion and Discussion}
The predictive value of various imaging parameters has previously been suggested in HCC. With the growing appreciation of tumor heterogeneity as a major obstacle to treatment response, more sophisticated image analysis algorithms are required. The complexity of such data analyses, especially considering multi-parametric multimodality imaging, requires computer aided techniques. We have presented a fully automatic framework to predict survival times of HCC patients. This approach based on fully convolutional and 3D convolutional neural networks outperformed state-of-the art handcrafted features, while still achieving the same diagnostic outcome as if human expert segmentations were provided. This work may have potential applications in HCC treatment planning.

\section{Acknowledgement}
This work was supported by the German Research Foundation (DFG) within the SFB-Initiative 824 (collaborative research center), ``Imaging for Selection, Monitoring and Individualization of Cancer Therapies" (SFB824, project C6) and the BMBF project Softwarecampus. We thank NVIDIA and Amazon AWS for granting GPU and computation support. 

\bibliographystyle{IEEEbib}
\bibliography{strings,refs}
\end{document}